\documentclass[12pt]{article}
\usepackage{amsmath,amssymb,graphicx}
\usepackage{hyperref,cancel,cite,float}
\usepackage[numbers]{natbib}

\title{Discrete vs. Continuous Trade-offs for Generative Models}
\author{Jathin Korrapati, Tanish Baranwal, Rahul Shah}
\date{University of California, Berkeley}
\begin{document}
%
\maketitle
\begin{abstract}
This work explores the theoretical and practical foundations of denoising diffusion probabilistic models (DDPMs) and score-based generative models, which leverage stochastic processes and Brownian motion to model complex data distributions. These models employ forward and reverse diffusion processes defined through stochastic differential equations (SDEs) to iteratively add and remove noise, enabling high-quality data generation. By analyzing the performance bounds of these models, we demonstrate how score estimation errors propagate through the reverse process and bound the total variation distance using discrete Girsanov transformations, Pinsker’s inequality, and the data processing inequality (DPI) for an information theoretic lens. 
\end{abstract}

\section{Introduction and Background}
\label{sec:intro}

%



\subsection{Background}
Before we start the introduction of our paper, we would like to provide some familiarity/introduction with core concepts relevant to our work, such as Brownian motion and score generative models. Readers familiar with these concepts may proceed directly to Section \ref{sec:derivation}.

\subsubsection{Background on denoising diffusion probabilistic modeling}
A denoising diffusion probabilistic model (DDPM) is a type of machine learning model that aims to estimate the underlying distribution of data to generate new similar data that could be found as generated from the same distribution ~\cite{ho2020denoising}. They are defined by a forward process which iteratively noises the data distribution, and a reverse process which recovers the original data distribution. The forward process transforms samples from some data distribution, i.e. $q$, into pure noise. The forward process is typically defined as:
\begin{equation}
    d\tilde{X} = -\tilde{X_t}dt + \sqrt{2}dB_t, \tilde{X_0} \sim{q} 
\end{equation}
The reverse process involves sampling the outputs of the forward process in order to transform the noise samples into the distribution sampled from $q$. The equation is:
\begin{equation}
    d\bar{X}^{\leftarrow}_{T-t} = \{\bar{X}^{\leftarrow}_{T-t} + 2\nabla \ln q_{T-t}(\bar{X}^{\leftarrow}_{T-t})\}dt + \sqrt{2}dB_t
\end{equation}
Where $\nabla \ln q_t$ is the score function for $q_t$ \cite{chen2022sampling}.

Since $q$ and thus $q_t$ are not explicitly known, the score function is typically estimated using a neural network based on samples. For this report, we will assume that there is a method to estimate the score with an error bound $\mathbb{E}[\|s_t-\nabla \ln q_t\|^2] \le \epsilon_{score}^2$, where $s_t$ is the estimated score. Similarly, since $q_T$ is not explicitly known, we can take advantage of the fact that $q_T \approx \gamma^d$ and initialize the algorithm at $X_0^\leftarrow \sim \gamma^d$, or from pure noise. The DDPM algorithm and subsequent diffusion models implement a discrete version of this score-matching algorithm. More precisely,

\begin{align}
    \bar{X}^\leftarrow_{T-(k+1)h} = e^h \bar{X}^\leftarrow_{kh} + 2s_{T-kh}(\bar{X}^\leftarrow_{T-kh})(e^{h}-1) + g, g\sim N(0, e^{2h}-1)
\end{align}
\subsubsection{Score Generative Models}
In particular, a score generative model does not try to model the data distribution, $q$, but instead, the score function, which is defined as the gradient of the log probability density using SDEs (stochastic differential equations):
\begin{equation}
    \nabla_x \log(q(x))
\end{equation}
In its de-noising process, it tries to estimate the score matching function instead of minimizing the $\mathcal{KL}$ divergence between the predicted distribution and $q$ ~\cite{song2020score}. 
Diffusion models, as shown in \cite{dhariwal2021diffusion}, surpass GANs in generating high-quality samples, particularly in image synthesis tasks.

\subsubsection{TVD and \texorpdfstring{$W_2$}{W2} Distances}
\label{subsubsec:tvd-w2}

When comparing probability distributions in the context of generative models, two key metrics are often employed: the Total Variation Distance (TVD) and the Wasserstein-2 ($W_2$) distance.

\paragraph{Total Variation Distance (TVD).}
For two probability measures $P$ and $Q$, the total variation distance is defined by
\begin{align}
TV(P, Q) \;=\; \sup_{A} \,\bigl|\,P(A) \;-\; Q(A)\bigr| \;=\; \tfrac{1}{2}\,\|P - Q\|_{1}.
\end{align}
This captures the maximum difference in the probabilities assigned by $P$ and $Q$ to any measurable event $A$; equivalently, the factor $\tfrac12$ ensures $TV(\cdot,\cdot)$ stays in $[0,1]$.

\paragraph{Wasserstein-2 Distance ($W_2$).}
Also called the 2nd-order Wasserstein or optimal transport distance, $W_2$ is defined for distributions $P$ and $Q$ over $\mathbb{R}^d$ as
\begin{align}
W_2(P, Q)
\;=\;
\biggl(\,\inf_{\pi\in\Gamma(P, Q)}\,\mathbb{E}_{(x, y)\sim\pi}
\bigl[\|\,x - y\,\|^2\bigr]\biggr)^{\!\tfrac12},
\end{align}
where $\Gamma(P, Q)$ is the set of joint measures on $\mathbb{R}^d \times \mathbb{R}^d$ whose marginals are $P$ and $Q$, respectively. Unlike TVD, $W_2$ incorporates the geometry of the underlying space and quantifies how much ``work'' or ``effort'' is required to transport one distribution into the other.

\subsubsection{Brownian Motion}
From the equations above, $(B_t)_{t\geq 0}$ is the standard motion for Brownian motion in $\mathbb{R}^d$. Brownian motion is defined as a stochastic process that models the random movement of particles in a fluid. It is often characterized by continuous random trajectories with independent and normally distributed increments over time. Score-generative models generate data by modeling a noising process and then learning to de-noise it (as described above) with the forward and reverse processes. These equations are often formalized as stochastic differential equations (SDEs) that are inspired by Brownian motion ~\cite{neurips2023}. Brownian motion describes a natural framework for modeling the continuous addition of noise to data, which, in turn, helps to generate and model complex data distributions. From the equations in the generative model, $b_t$ represents the drift term that is approximated using the score function, and $\sigma_t$ controls the rate of diffusion.

\subsection{Introduction}
\label{sec:derivation}
Practically, score generative models have had considerable success, starting with the DDPM \cite{ho2020denoising} paper in 2020. However, until \cite{chen2022sampling}, there wasn't a definitive bound on how well score generative models capture the true score matching process described earlier.

For $P_T$ being the law of the SGM reverse process initialized at $\gamma^d$, the standard isotropic gaussian and using the estimated score, and $Q_T$ being the law of the reverse OU process initialized at the fully noised point $q_T$ and with score $\nabla \log q_t$, we want to bound the difference between these two probability distributions. This can be done using total variation distance.

To bound Total Variation Distance, we start with the data processing inequality (DPI)\cite{thomasandcover}:
\begin{align}
TV(P_T, Q_T) 
&\leq TV(P_T, P_T^{q_T}) + TV(P_T^{Q_T}, Q_T^{\leftarrow})
\\
&\leq TV(q_T, \gamma^d) + TV(P_T^{Q_T}, Q_T^{\leftarrow})
\end{align}

Pinsker's inequality\cite{pinsker} says:
\begin{align}\label{eq:pinsker}
TV(P_T^{Q_T}, Q_T^{\leftarrow})^2 
&\leq {\frac{1}{2} \mathcal{KL}(P_T^{Q_T} \| Q_T^{\leftarrow})}.
\end{align}

\hrulefill

We have that $P_T^{Q_T}=P(X)$ and $Q_T^{\leftarrow} = Q(X)$, where:
\[
P := \text{law of SGM algo initialized at fully noised point } q_T, \text{ law of $P_t^{Q_t}$, with } S_{kh}{\theta}.
\]
\[
Q := \text{law of the reverse process with } \nabla \log q_{kh} \text{ (true score)}
\]

By Discrete Girsanov and the $\mathcal{KL}$ formula proved in \ref{girsanovproof}:
\begin{align}
\mathcal{KL}\left(P(X) \| Q(X)\right) 
&= \mathbb{E}_{P(x)} \left[\frac14 \sum_{k=0}^{N-1} {h} \|S_{kh}(\theta) - \nabla \log q_{kh}\|_2^2 \right]
\\
&= O\left(T \epsilon_{\text{score}}^2\right)
\end{align}

as
$\|S_{kh}(\theta) - \nabla \log q_{kh}\|_2 
\leq \epsilon_{\text{score}}
\implies
\|S_{kh}(\theta) - \nabla \log q_{kh}\|_2^2 
\leq \epsilon_{\text{score}}^2.$

Thus, taking the $\sqrt{\cdot}$ of both sides of \eqref{eq:pinsker}, we get that:
\begin{align}
TV(P_T^{Q_T}, Q_T^{\leftarrow}) \leq O\left(\sqrt{T} \cdot \epsilon_{\text{score}}\right).
\end{align}

For the forward OU process, we know that it's stationary distribution is $\gamma^d$, moreover it converges to $\gamma^d$ exponentially fast in time\cite{bakry2014analysis}:

\begin{align}
    W_2(q_t, \gamma^d) \le e^{-t}W_2(q, \gamma^d)
\end{align}

Where $q_T = \text{law}(X_t)$.

Talgrand's inequality\cite{tao2009talagrand} for the Gaussian measure $\gamma^d$ states that $W_2(q, \gamma^d) \le \sqrt{2KL(q\|\gamma^d)}$, and $W_2(q_T, \gamma^d) \le e^{-T}\sqrt{2KL(q\|\gamma^d)}$.

Since the reverse process also has a drift that is close to the OU drift, as $T\rightarrow\infty, q_{T-\tau}$ approaches $\gamma^d$, making $\nabla\ln q_{T-\tau}(x) \approx -x$. Thus, the reverse process closely resembles another stable OU-type process. Since the initial distributions differ by at most $W_2(q_T, \gamma^d)$, $W_2(q, p_T) \lesssim W_2(q+T, \gamma^d)$.

Thus, we can show that $W_2(q, p_T) \le Ce^{-T} \sqrt{\mathcal{KL}(q\|\gamma^d)}$. Working under the Gaussian measure we can show that $TV(q, p_T) \le W_2(q, p_T)$, and thus:

\begin{align}
    TV(q, p_T) \le O\left(e^{-T} \sqrt{\mathcal{KL}(q\|\gamma^d)}\right)
\end{align}

Which is what we want to bound.

Thus, combining this with the previous result, we find the bound for the overall error of the score-matching algorithm and the true reverse process.

\begin{align}
    TV(P_T, Q_T) \le O\left(\sqrt{T}\epsilon_{score} + e^{-T}\sqrt{\mathcal{KL}(q\|\gamma^d)}\right)
\end{align}
\subsubsection{Analysis Approach}

We directly analyze the discrete definition of score matching models by first deriving a discrete analog to Girsanov's theorem \cite{girsanov}, and then bounding the error in the Denoising Diffusion Probabilistic Models framework (DDPM) \cite{song2020score} using discrete Girsanov to bound the error due to the score estimation error, and bounding the initialization error due to initializing at an isotropic gaussian.

\subsubsection{Statement of Girsanov's Theorem}
Here we introduce the concept of Girsanov's Theorem, which is core to our paper. Girsanov's theorem is a result in stochastic calculus that describes how the measure associated with a stochastic process changes when the drift term of a Brownian motion is altered. At its core, Girsanov's Theorem provides a framework for relating two probability measures under a change of drift, which is particularly useful in the context of stochastic differential equations (SDEs). The relationship between stochastic interpolants and Girsanov's theorem is particularly useful for understanding changes in drift \cite{albergo2023stochastic}. Girsanov's Theorem states that there exists a new measure \( \mathbb{Q} \), absolutely continuous with respect to \( \mathbb{P} \), under which the process \( X_t \) becomes a standard Brownian motion. The relationship between the two measures \( \mathbb{P} \) and \( \mathbb{Q} \) is given by the Radon-Nikodym derivative:

\[
\frac{d\mathbb{Q}}{d\mathbb{P}} \Bigg|_{\mathcal{F}_t} = \exp\left( -\int_0^t \mu_s \, dW_s - \frac{1}{2} \int_0^t \|\mu_s\|^2 \, ds \right),
\]
where \( \mathcal{F}_t \) is the natural filtration up to time \( t \), \( \mu_s \) is the drift term, and \( W_t \) is the original Brownian motion under measure \( \mathbb{P} \).

\subsection{Proof of Girsanov's Discrete theorem}
\label{girsanovproof}

\begin{enumerate}
\item[(1)] $x_{(k+1) h}
=x_{kh}+h \cdot b_{k h}\left(x_{k h}\right)+\sqrt{2 h} g_{kh}$.

\item[(2)] $x_{(k+1) h}
=x_{k h}+h b_{k h}^{\prime}\left(x_{k h}\right)+\sqrt{2 h} \tilde{g}_{k h}$, \\
$b_{k h}\left(x_{k h}\right), b_{k h}^{\prime}\left(x_{k h}\right)$ are any two functions $\tilde{g}_{kh}, g_{k h} \sim N\left(0, I_d\right)$
\end{enumerate}

Say we have trajectory $X=\left(x_0, x_h, x_{2 h}, \cdots x_{N h}\right)$

Under process (1), the likelihood of X is:
$$
P(x) = \prod_{k=0}^{N-1} \exp\left(-\frac{\|x_{(k + 1)h} - (x_{kh} +hb_{kh}(x_{kh}))\|^2}{4h}\right)
$$

Under process (2), the likelihood of X is:
$$
Q(x) = \prod_{k=0}^{N-1} \exp\left(-\frac{\|x_{(k + 1)h} - (x_{kh} +hb_{kh}^\prime(x_{kh}))\|^2}{4h}\right)
$$

We aim to quantify the difference between the distributions $P(X)$ and $Q(X)$. The most natural way for us to do so is via using the $\mathcal{KL}$ Divergence:
\begin{align}
    \mathcal{\mathcal{KL}}(P(X) \| Q(X)) = \mathbb{E}_{P(X)}\left[\log \frac{P(x)}{Q(x)}\right]
\end{align}

Define \(b_{kh} = b_{kh}(X_{kh})\) and \(b_{kh}^\prime = b_{kh}^\prime(X_{kh})\).

Define \(\Delta_k = X_{{(k+1)}h} - X_{kh}\).

Define \(\Delta_b = b_{kh} - b'_{kh}\)

\begin{align}
    \frac{P(X)}{Q(X)} = \prod_{k=0}^{N-1} \exp\left(-\frac{1}{4h} \|\Delta_k - h b_{kh}\|^2 - \|\Delta_k-kb_{kh}^\prime \|^2\right)
\end{align}

\begin{align}
d_1\triangleq\| \Delta_k - h b_{kh}\|^2 
&= \|\Delta_k\|^2 - 2h \langle \Delta_k, b_{kh}\rangle + h^2 \|b_{kh}\|^2
\\
d_2\triangleq\| \Delta_k - h b_{kh}^\prime\|^2 
&= \|\Delta_k\|^2 - 2h \langle \Delta_k, b_{kh}^\prime\rangle + h^2 \|b_{kh}^\prime\|^2
\\
d_1 - d_2
= h^2\|b_{kh}\|^2 
&- h^2 \|b_{kh}^\prime\|^2
-2h\langle \Delta_k, b_{kh}\rangle - \langle \Delta_k, b_{kh}^\prime\rangle
\end{align}

Thus:
\begin{align}
\frac{P(X)}{Q(X)} = \prod_{k=0}^{N-1} \exp\left(\frac{-1}{4h} \left[h^2 \|b_{kh}\|^2 - h^2 \|b_{kh}^\prime\|^2 - 2h \langle \Delta_k, \Delta_b\rangle \right]\right)
\end{align}

Let \(\Delta_k = h b_{kh}^\prime + \sqrt{2h} g_{kh}\).\quad = \(X_{{(k+1)}h} - X_{kh} 
\ \boxed{\ast}\text{ under process (2)}\):

\begin{align}
h^2\|b_{kh}\|^2
&
- h^2 \|b_{kh}^\prime\|^2
-2h\langle h b_{kh}^\prime, 
b_{kh} - b_{kh}^\prime\rangle
\\
&h^2\langle b_{kh}+b_{kh}^\prime-2b_{kh}^\prime, 
b_{kh} - b_{kh}^\prime\rangle
\\
&=h^2\|b_{kh}-b_{kh}^\prime\|
\\
&= h^2\|\Delta_b\|
\end{align}

\begin{align}
\frac{P(X)}{Q(X)} 
    &= \prod_{k=0}^{N-1} \exp\left(-\frac{1}{4h} \left[h^2 \|\Delta_b\|^2 - 2h\sqrt{2} \langle \sqrt{h} g_{kh}, \Delta_b\rangle \right]\right)
    \\
    &= \exp\left(-\frac{1}{4} \sum_{k=0}^{N-1} h \|\Delta_b\|^2 
    + \frac{\sqrt{2}}2 \sum_{k=0}^{N-1} 
    \langle \sqrt{h} \tilde{g}_{kh}, 
    \Delta_b\rangle \right)
\end{align}
setting processes (1), (2) equal, $X_{kh} + hb_{kh} + \sqrt{2h} g_{kh} = X_{kh} + hb_{kh}^\prime + \sqrt{2h} \tilde{g}_{kh}$.
\begin{align}
    \sqrt{h}\tilde{g}_{hh} = \frac{1}{\sqrt{2}}h(\Delta_b)+\sqrt{h}g_{kh}
\end{align}

\begin{align}
    \frac{P(X)}{Q(X)} &= \exp\left(\sum^{N-1}_{k=0}\left[-\frac{1}{4}h \| \Delta_b\|^2 + \frac{1}{\sqrt{2}}\langle \frac{1}{\sqrt{2}}h\Delta_b, \Delta_b \rangle + \frac{1}{\sqrt{2}}\langle \sqrt{h}g_{kh}, \Delta_b\rangle\right]\right) \\
    &= \exp\left(-\frac{1}{4}\sum^{N-1}_{k=0}h \| \Delta_b\|^2 + \frac{1}{\sqrt{2}}\sum^{N-1}_{k=0} \langle \sqrt{h} g_{kh}, \Delta_b \rangle \right)
\end{align}

\hrulefill

Note that Discrete Girsanov is usually written as the reciprocal $\frac{Q(X)}{P(X)}$. \\
We reciprocate later between (\ref{eq:dP-dQ}) and (\ref{eq:dQ-dP}).

\begin{align}
    \mathcal{KL}(P(X) \| Q(X)) 
    &= \mathbb{E}\left[
    \log\frac{P(X)}{Q(X)}\right] \\
    &= \mathbb{E}\left[\frac{1}{4}\sum^{N-1}_{k=0}h \| b_{kh}-b'_{kh}\|^2
    +\frac1{\sqrt2}\sum_{k=0}^{N-1} \sum_{h=0}^{N-1}\cancelto{0}{\langle \sqrt{h}g_{kh}, b_{kh}-b_{kh}^\prime\rangle}
    \quad
    \right]
    \\
    \mathcal{KL}(P(X) \| Q(X)) &= 
    \mathbb{E}_{P(X)}\left[\frac{1}{4}\sum^{N-1}_{k=0}h \| b_{kh}-b'_{kh}\|^2\right]
\end{align}

This discrete version of Girsanov's theorem offers a pivotal link between continuous stochastic calculus and discrete-time simulations -- it allows us to rigorously bound how deviations in the drift at each discrete timestep accumulate into global errors, ensuring stability and theoretical guarantees -- which is essential for algorithms like DDPMs.

From an information-theoretic viewpoint, the discrete Girsanov bound indicates how many “bits” are required to distinguish or measure a drift mismatch at each timestep, thereby ensuring that the global KL divergence across the entire process remains bounded. By linking drift differences to this bit cost, we gain a clearer picture of how small local changes -- in drift estimates -- can accumulate into large divergences if not carefully controlled.

\begin{figure}[H]
    \centering
    \includegraphics[width=0.85\linewidth]{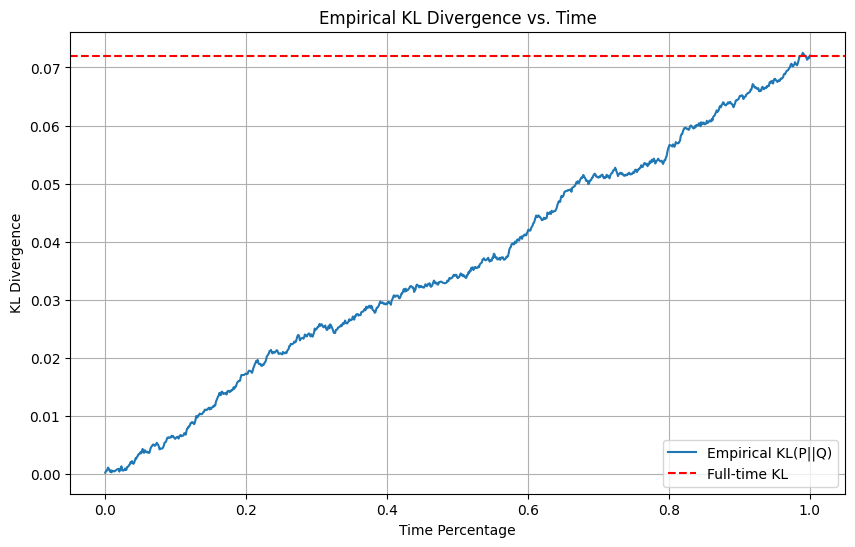}
    \caption{Here we see that the $\mathcal{KL}$ divergence is clearly linear over time}
    \label{fig:empirical-kl}
\end{figure}

\hrulefill

Typically, Girsanov's says that $P$-Brownian motion $B_t$ can be modified to $Q$-Brownian motion by suitably changing the drift. Specifically: $$\tilde{B}_t = B_t + \int_0^t \gamma_s \, \mathrm ds$$ for drift $\gamma_s$.

Consider a process $x_t$ evolving under two measures $P$ and $Q$. Under $P$, we have the equivalent SDE:
\begin{align}
    dx_t = b_t\,dt + \sqrt{2}\,dB_t,
\end{align}

where $B_t$ is a Brownian motion under $P$ and $b_t$ is the drift under $P$.

Under $Q$, the same process evolves as:
\begin{align}
dx_t = b_t^\prime\,dt + \sqrt{2}\,d\tilde{B}_t,
\end{align}

where $\tilde{B}_t$ is a Brownian motion under $Q$ and $b_t^\prime$ is the drift under $Q$.

Define
\begin{align}\label{eq:gamma-def}
\gamma_t = \frac{b_t - b_t^\prime}{\sqrt{2}}.
\end{align}
Then we can write
\begin{align}
d\tilde{B}_t = dB_t + \gamma_t\,dt.
\end{align}

Girsanov's theorem says that the Radon–Nikodym derivative\cite{radon_nikodym_derivative} gives the change of measure from $P$ to $Q$:
\begin{align}
\label{eq:dQ-dP}
\frac{dQ}{dP} = \exp\left(-\int_0^T \gamma_s\,dB_s - \tfrac{1}{2}\int_0^T \gamma_s^2\,ds\right).
\end{align}

Inverting this relationship and sub'ing in eq. (\ref{eq:gamma-def}) for $\gamma_s$, we get:
\begin{align}\label{eq:dP-dQ}
\frac{dP}{dQ} 
&= \exp\left(\frac{1}{\sqrt{2}}\int_0^T (b_s - b_s^\prime)\,d{B}_s + \frac{1}{4}\int_0^T (b_s - b_s^\prime)^2\,ds\right)
\\
&= \exp\left(\frac1{\sqrt2}\sum_{k=0}^{N-1}\langle \sqrt{h}g_{kh}, b_t - b_t^\prime\rangle + \frac14\sum_{k=0}^{N-1}\|b_t - b_t^\prime\|^2 \cdot h\right).
\end{align}

This is precisely the relationship we aimed to prove.

\begin{figure}[H]
    \centering
    \includegraphics[width=0.8\linewidth]{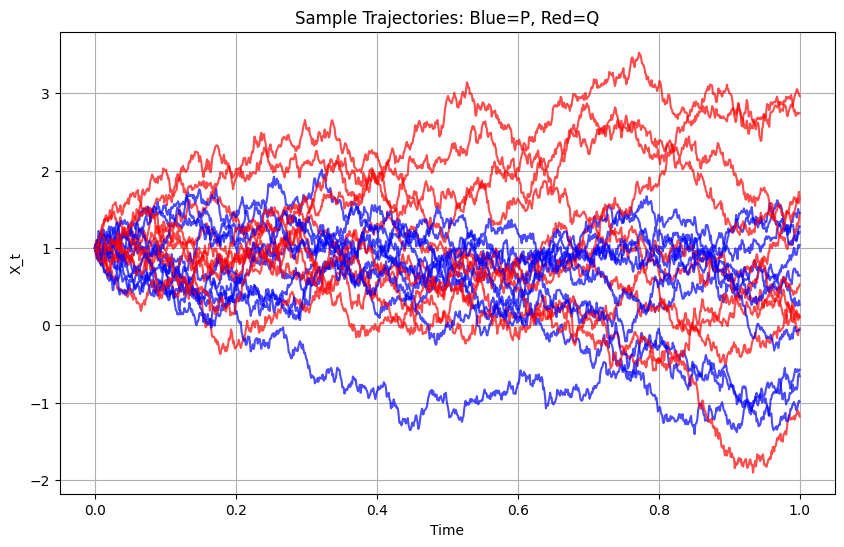}
    \caption{Sample Trajectories}
    \label{fig:sample-trajectories}
\end{figure}

Note that when strongly convex, all converge to the same point so there's an error correction aspect where you can take too large or too small which is inverse poly.

\subsection{Implementation \& Results}
For our experiment, we simulate an OU process with the forward and reverse processes with what we expect and what we simulate, which we use to show the significance of the discrete Girsanov Theorem here.
\begin{enumerate}
\item[1.] Define a true forward OU process that evolves from some initial data. We generate data from the distribution using the function \(y = x\).
\item[2.] From the noising equations from above, we create a noised version of the data by simulating the forward process to time \(T\).
\item[3.] Define two reverse processes:
\begin{itemize}
    \item The correct reverse process (what we expect from ground truth).
   \item The estimated reverse process (with an estimated score that has some error).
\end{itemize}
\end{enumerate}

We will then:
\begin{itemize}
  \item Simulate sample paths from both the correct and approximate reverse processes starting from the final noised distribution.  
  \item Compare the denoised outputs \(\tilde{y}\) to the original data \(y\).  
  \item Plot the original data, the noised version, and the denoised versions.  
  \item Plot the errors and estimate the $\mathcal{KL}$ divergence between the two distributions using the discrete Girsanov formula derived in the paper.
\end{itemize}
\begin{figure}[H]
    \centering
    \includegraphics[width=0.85\linewidth]{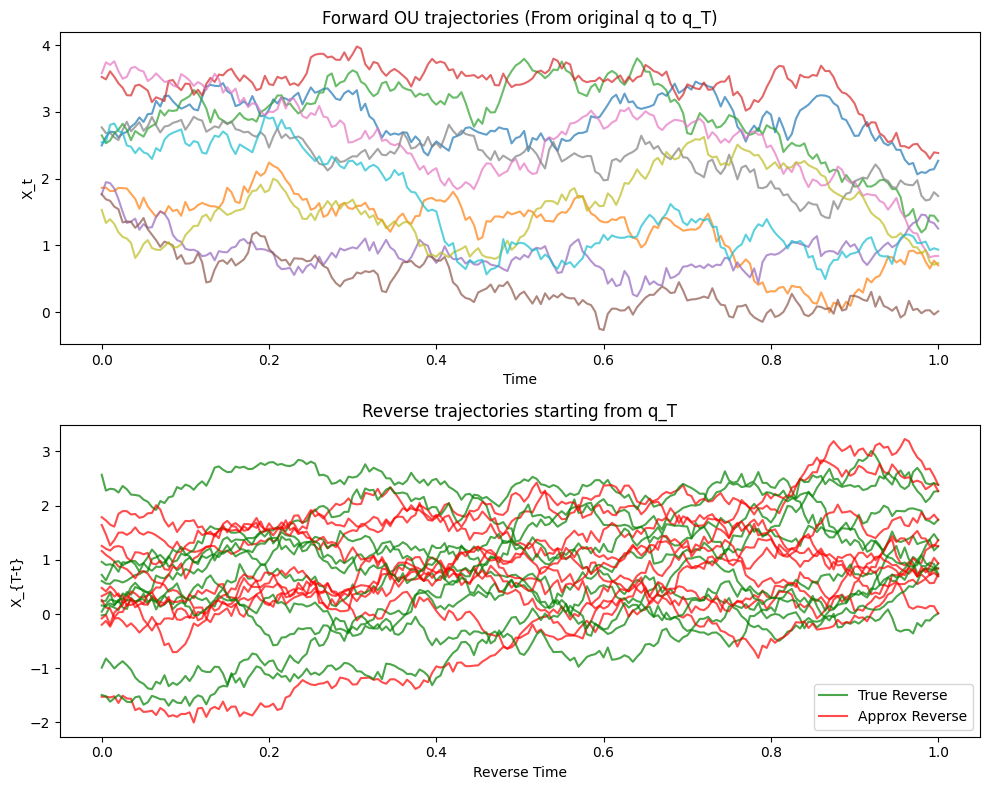}
    \caption{Forward and Reverse Trajectories from original $q$ to $q_T$}
    \label{fig:trajs}
\end{figure}

\begin{figure}[H]
    \centering
    \includegraphics[width=0.85\linewidth]{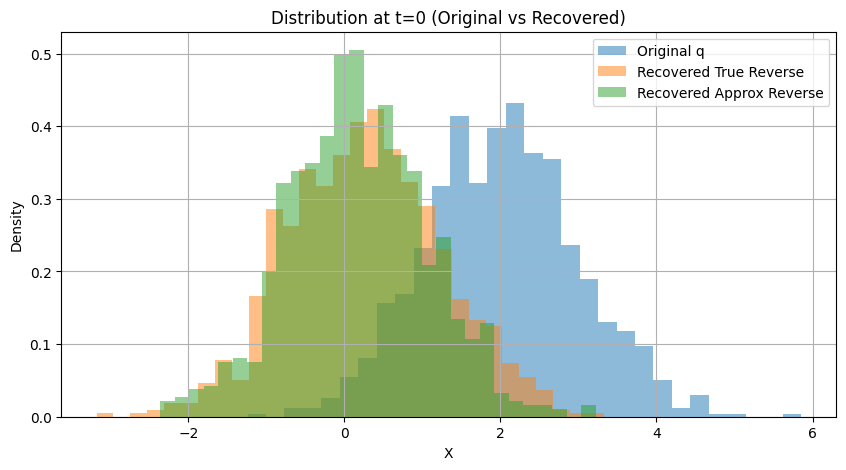}
    \caption{Original Distribution vs. Recovered Distribution of Data}
    \label{fig:reconstruction}
\end{figure}

\begin{figure}[H]
    \centering
    \includegraphics[width=0.85\linewidth]{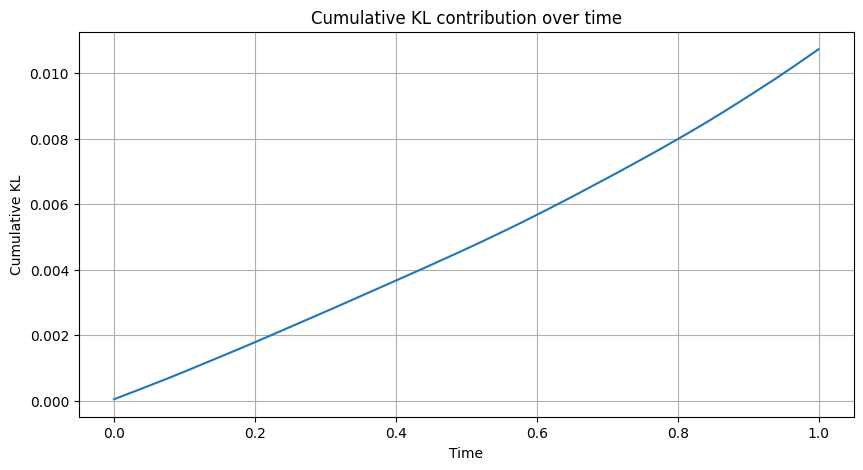}
    \caption{Here we see how drift mismatches contribute to global errors via the Cumulative $\mathcal{KL}$ Divergence Plot, as demonstrated by eq. \eqref{eq:dP-dQ}}
    \label{fig:cum_kl}
\end{figure}

\begin{figure}[H]
    \centering
    \includegraphics[width=0.85\linewidth]{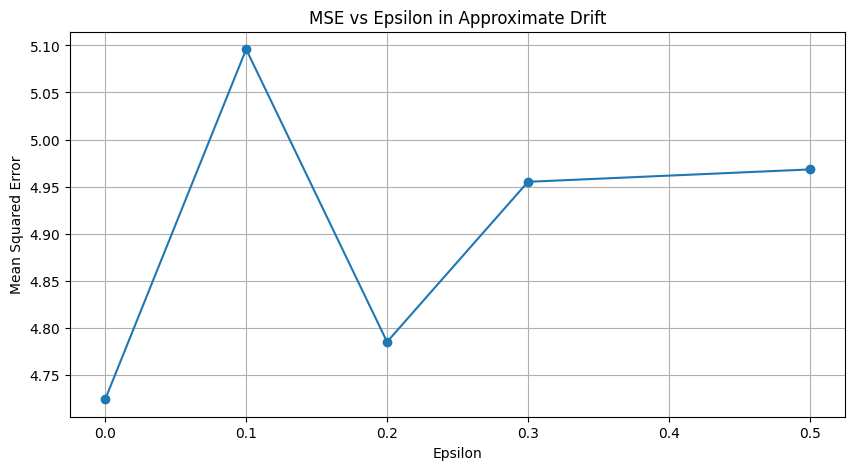}
    \caption{Brownian Motion Approximation Drift across time with $\mathcal{KL}$ Divergence}
    \label{fig:mse}
\end{figure}

These results show the significance of the data distribution we generated from the process above -- specifically they highlight the accuracy of the discrete Girsanov estimates in recovering the original data distribution -- where we first plot the original and reverse trajectories (as seen in Figure \ref{fig:trajs}). We see using the discrete Girsanov estimates that we are able to estimate the original $q$ distribution well, as seen in Figure \ref{fig:reconstruction}. Figures \ref{fig:cum_kl} and \ref{fig:mse} display the errors for the $\mathcal{KL}$ divergence and Brownian motion, which both represent low errors/KL, meaning our discrete Girsanov representation estimates the distribution well.


\subsection{Benefits \& Drawbacks between Discrete and Continuous Methods}
\begin{table}[H]
    \centering
    \begin{tabular}{|p{6cm}|p{6cm}|}
        \hline
        \textbf{Discrete Approach} & \textbf{Continuous Approach} \\ 
        \hline
        Time is partitioned into a finite number of steps, $N$, and the process evolves step-by-step. & Time is treated as a continuum, with the process defined by SDEs over an infinite or uncountably many time points. \\ 
        \hline
        Approximates the underlying continuous dynamics through numerical schemes (e.g., Euler–Maruyama), which may introduce discretization error. & Directly defined by stochastic differential equations (SDEs) or continuous-time Markov processes, capturing exact continuous behavior. \\ 
        \hline
        Often more tractable computationally for simulation and easier to code (due to iterative updates). & Elegant and closely tied to theoretical tools in stochastic calculus, at the cost of simulation. \\ 
        \hline
        Analysis is usually simpler but may lose some fine-grained properties of the continuous process due to discretization. & Offers richer analytical insights and exact theoretical results (e.g., closed-form solutions for OU processes, use of Girsanov's), but more difficult to directly simulate. \\ 
        \hline
        Complexity often scales linearly with the number of discrete steps $N$; achieving low error may require large $N$. & Continuous-time models do not have a discrete “step” parameter, but achieving small approximation errors may require specialized solvers or refined discretizations, still leading to higher computational costs. \\ 
        \hline
        Useful for constructing algorithms like DDPMs or discrete-time MCMC methods, where discrete iteration is intrinsic. & Useful for theoretical guarantees, deriving continuous-time bounds, and leveraging tools from stochastic calculus (e.g., continuous Girsanov transformations). \\ 
        \hline
    \end{tabular}
    \caption{Comparison of Discrete \& Continuous Stochastic Models}
    \label{tab:comparison_discrete_continuous}
\end{table}

\newpage
In many cases, discrete approximations can serve as a bridge to their continuous counterparts, allowing insights and results from discrete-time settings to be refined and extended into continuous-time frameworks. Discrete versions are often easier to implement computationally, providing practitioners with finer control over numerical stability, step-size selection, and algorithmic design. This level of control can facilitate rapid experimentation, debugging, and the integration of techniques like adaptive step sizing or variance reduction strategies that might be less straightforward in purely continuous formulations. Conversely, continuous-time models, while theoretically elegant and rich in analytical tools, may be more challenging to simulate directly, often requiring specialized solvers or sophisticated approximations. Thus, the discrete analogs not only approximate their continuous counterparts but can also offer practical advantages and a more flexible platform for testing, refinement, and exploration before passing to the continuous case -- in the limit.

While Brownian motion is inherently continuous, score matching is a discrete process (due to DDPM being a discrete algorithm) and as such, discretization works very well giving not only a more tractable solution but also a more efficient one. Thus we are able to tackle a series of problems that we otherwise could not.

\subsection{Conclusion}
In this paper, we explored several of the computational trade-offs between discrete and continuous methods, such as Denoising Diffusion Probabilistic Models, highlighting that while continuous methods are theoretically grounded, their computational complexity and challenges in modeling often limit their practical application. By leveraging Girsanov's Theorem to bound our results, we were able to establish rigorous mathematical foundations that enable more efficient simulation generation. This approach not only provides valuable insights into the relationship between discrete and continuous methods but also lays the groundwork for future research by offering a robust framework for bounding and refining model predictions in complex systems.

\bibliographystyle{IEEEtran}
\bibliography{references}

\begin{thebibliography}{10}
\providecommand{\url}[1]{#1}
\csname url@samestyle\endcsname
\providecommand{\newblock}{\relax}
\providecommand{\bibinfo}[2]{#2}
\providecommand{\BIBentrySTDinterwordspacing}{\spaceskip=0pt\relax}
\providecommand{\BIBentryALTinterwordstretchfactor}{4}
\providecommand{\BIBentryALTinterwordspacing}{\spaceskip=\fontdimen2\font plus
\BIBentryALTinterwordstretchfactor\fontdimen3\font minus
  \fontdimen4\font\relax}
\providecommand{\BIBforeignlanguage}[2]{{%
\expandafter\ifx\csname l@#1\endcsname\relax
\typeout{** WARNING: IEEEtran.bst: No hyphenation pattern has been}%
\typeout{** loaded for the language `#1'. Using the pattern for}%
\typeout{** the default language instead.}%
\else
\language=\csname l@#1\endcsname
\fi
#2}}
\providecommand{\BIBdecl}{\relax}
\BIBdecl

\bibitem{ho2020denoising}
\BIBentryALTinterwordspacing
J.~Ho, A.~Jain, and P.~Abbeel, ``Denoising diffusion probabilistic models,''
  \emph{arXiv preprint arXiv:2006.11239}, 2020. [Online]. Available:
  \url{https://arxiv.org/abs/2006.11239}
\BIBentrySTDinterwordspacing

\bibitem{chen2022sampling}
\BIBentryALTinterwordspacing
S.~Chen, S.~Chewi, J.~Li, Y.~Li, A.~Salim, and A.~R. Zhang, ``Sampling is as
  easy as learning the score: Theory for diffusion models with minimal data
  assumptions,'' \emph{arXiv preprint arXiv:2209.11215}, 2022. [Online].
  Available: \url{https://arxiv.org/abs/2209.11215}
\BIBentrySTDinterwordspacing

\bibitem{song2020score}
\BIBentryALTinterwordspacing
Y.~Song, J.~Sohl-Dickstein, D.~P. Kingma, A.~Kumar, S.~Ermon, and B.~Poole,
  ``Score-based generative modeling through stochastic differential
  equations,'' \emph{arXiv preprint arXiv:2011.13456}, 2020. [Online].
  Available: \url{https://arxiv.org/pdf/2011.13456}
\BIBentrySTDinterwordspacing

\bibitem{dhariwal2021diffusion}
P.~Dhariwal and A.~Nichol, ``Diffusion models beat gans on image synthesis,''
  2021.

\bibitem{neurips2023}
\BIBentryALTinterwordspacing
E.~Yoon, K.~Park, S.~Kim, and S.~Lim, ``Score-based generative models with
  lévy processes,'' \emph{NeurIPS Conference Proceedings}, 2023. [Online].
  Available:
  \url{https://papers.neurips.cc/paper_files/paper/2023/file/8011b23e1dc3f57e1b6211ccad498919-Paper-Conference.pdf}
\BIBentrySTDinterwordspacing

\bibitem{thomasandcover}
T.~M. Cover and J.~A. Thomas, \emph{Elements of Information Theory},
  2nd~ed.\hskip 1em plus 0.5em minus 0.4em\relax Hoboken, NJ, USA:
  Wiley-Interscience, 2006.

\bibitem{pinsker}
M.~S. Pinsker, ``Information and information stability of random variables and
  processes,'' \emph{Soviet Math. Dokl.}, vol.~4, pp. 537--541, 1964.

\bibitem{bakry2014analysis}
\BIBentryALTinterwordspacing
D.~Bakry, I.~Gentil, and M.~Ledoux, \emph{Analysis and Geometry of Markov
  Diffusion Operators}, ser. Grundlehren der mathematischen
  Wissenschaften.\hskip 1em plus 0.5em minus 0.4em\relax Cham: Springer, 2014,
  vol. 348. [Online]. Available:
  \url{https://link.springer.com/book/10.1007/978-3-319-00227-9}
\BIBentrySTDinterwordspacing

\bibitem{tao2009talagrand}
\BIBentryALTinterwordspacing
T.~Tao, ``Talagrand's concentration inequality,'' 2009, accessed: 2024-04-27.
  [Online]. Available:
  \url{https://terrytao.wordpress.com/2009/06/09/talagrands-concentration-inequality/}
\BIBentrySTDinterwordspacing

\bibitem{girsanov}
\BIBentryALTinterwordspacing
I.~V. Girsanov, ``On transforming a certain class of stochastic processes by
  absolutely continuous substitution of measures,'' \emph{Theory of Probability
  \& Its Applications}, vol.~5, no.~3, pp. 285--301, 1960. [Online]. Available:
  \url{https://doi.org/10.1137/1105027}
\BIBentrySTDinterwordspacing

\bibitem{albergo2023stochastic}
M.~S. Albergo, N.~M. Boffi, and E.~Vanden-Eijnden, ``Stochastic interpolants: A
  unifying framework for flows and diffusions,'' 2023.

\bibitem{radon_nikodym_derivative}
\BIBentryALTinterwordspacing
T.~Rowland, ``Radon–nikodym derivative,'' \emph{MathWorld--A Wolfram Web
  Resource}, 2023. [Online]. Available:
  \url{https://mathworld.wolfram.com/Radon-NikodymDerivative.html}
\BIBentrySTDinterwordspacing

\end{thebibliography}

\end{document}